\documentclass[journal]{IEEEtran}
\usepackage{dsfont}
\usepackage{bbm}
\usepackage{array}
\makeatletter
\newcommand{\thickhline}{%
    \noalign {\ifnum 0=`}\fi \hrule height 1.5pt
    \futurelet \reserved@a \@xhline
}
\newcolumntype{"}{@{\hskip\tabcolsep\vrule width 1pt\hskip\tabcolsep}}
\makeatother
\usepackage{amsfonts}
\usepackage{color}
\usepackage{multicol}
\usepackage{booktabs}
\usepackage{graphicx}
\usepackage{setspace}
\usepackage{amsmath}
\usepackage{amssymb}
\usepackage{cite}
\usepackage[pagebackref=true,breaklinks=true,letterpaper=false,colorlinks,citecolor=blue,bookmarks=true]{hyperref}
\makeatletter
\newcommand{\rmnum}[1]{\romannumeral #1}
\newcommand{\Rmnum}[1]{\expandafter\@slowromancap\romannumeral #1@}
\makeatother
\ifCLASSINFOpdf
\else
\fi
\begin{document}
%
\title{Vision-to-Language Tasks Based on Attributes and Attention Mechanism}
%
%
%

\author{Xuelong Li,~\IEEEmembership{Fellow,~IEEE,}
        Aihong Yuan,
        and~Xiaoqiang Lu,~\IEEEmembership{Senior Member,~IEEE}
\thanks{This work was supported in part by the National Natural Science Foundation of China under Grant no. 61772510 and in part by the Young Top-Notch Talent Program of Chinese Academy of Sciences under Grant no. QYZDB-SSWJSC015. \emph{(Corresponding author: Xiaoqiang Lu)}}
\thanks{Xuelong Li  is with School of Computer Science and Center for OPTical IMagery Analysis and Learning (OPTIMAL), Northwestern Polytechnical University, Xi'an 710072, P.R. China.}
\thanks{Aihong Yuan is with the Key Laboratory of Spectral Imaging Technology CAS, Xi'an Institute of Optics and Precision Mechanics, Chinese Academy of Sciences, Xi'an 710119, Shaanxi, P. R. China,  and also with the University of Chinese Academy of Sciences, Beijing 100049, China.}
\thanks{Xiaoqiang Lu is with the Key Laboratory of Spectral Imaging Technology CAS, Xi'an Institute of Optics and Precision Mechanics, Chinese Academy of Sciences, Xi'an 710119, Shaanxi, P. R. China.}
\thanks{Color versions of one or more of the figures in this paper are available online at http://ieeexplore.ieee.org.}
\thanks{20XX IEEE. Personal use of this material is permitted. Permission from IEEE must be obtained for all other uses, in any current or future media, including reprinting/republishing this material for advertising or promotional purposes, creating new collective works, for resale or redistribution to servers or lists, or reuse of any copyrighted component of this work in other works.}
}

%
%

\markboth{IEEE TRANSACTIONS ON CYBERNETICS,~2019}%
{Shell \MakeLowercase{\textit{et al.}}: 3G Structure for Image Caption Generation}
%



\maketitle

\begin{abstract}
  Vision-to-Language tasks aim to integrate computer vision and natural language processing together, which has attracted the attention of many researchers. For typical approaches, they encode image into feature representations and decode it into natural language sentences. While they neglect high-level semantic concepts and subtle relationships between image regions and natural language elements. To make full use of these information, this paper attempt to exploit the text-guided attention and semantic-guided attention to find the more correlated spatial information and reduce the semantic gap between vision and language. Our method includes two level attention networks. One is the text-guided attention network which is used to select the text-related regions. The other is semantic-guided attention network which is used to highlight the concept-related regions and the region-related concepts. At last, all these information are incorporated to generate captions or answers. Practically, image captioning and visual question answering experiments have been carried out, and the experimental results have shown the excellent performance of the proposed approach.
\end{abstract}


%
\IEEEpeerreviewmaketitle

\section{Introduction}
%
%
%
%
\IEEEPARstart{V}ision-to-Language (V2L) tasks aim to integrate natural language processing and computer vision together. Typical V2L tasks are image captioning \cite{ATT2U, DBLP:conf/aaai/LiuMSY17, DBLP:conf/cits/QuLTL16, Binqiang}, visual question answering (VQA) \cite{DBLP:conf/icml/XiongMS16, DBLP:conf/cvpr/ShihSH16, TGIF-QA} and video description \cite{DBLP:conf/cvpr/DonahueHGRVDS15, HVC, TDVD, DBLP:conf/ijcai/LiZL17, DBLP:conf/mm/ZhaoLL17}. Recently, due to the advent of artificial intelligence (AI), V2L tasks have attracted extensive attention. Practically, V2L tasks enable many important applications, including early childhood education, human-robot interaction, visually impaired people assistance and so on.

\par Many recent approaches for V2L tasks have achieved a lot of gratifying results through combining Convolutional Neural Networks (CNNs) and Recurrent Neural networks (RNNs) for image encoding and text generating, respectively \cite{DBLP:conf/cvpr/KarpathyL15, DBLP:conf/cvpr/VinyalsTBE15, DBLP:journals/corr/MaoXYWY14a, DBLP:conf/nips/RenKZ15}. Concretely, a CNN pre-trained on ImageNet \cite{DBLP:journals/ijcv/RussakovskyDSKS15} is used to extract global image feature while a RNN is used to encode the language information. Most of the recently approaches are belong to the ``CNN-RNN'' paradigm and these approaches have attained some promising results, further improvements should be got over some limitations.

\subsection{Motivation and Overview}

\begin{figure}[t]
  \centering
  \includegraphics[width=0.90\linewidth]{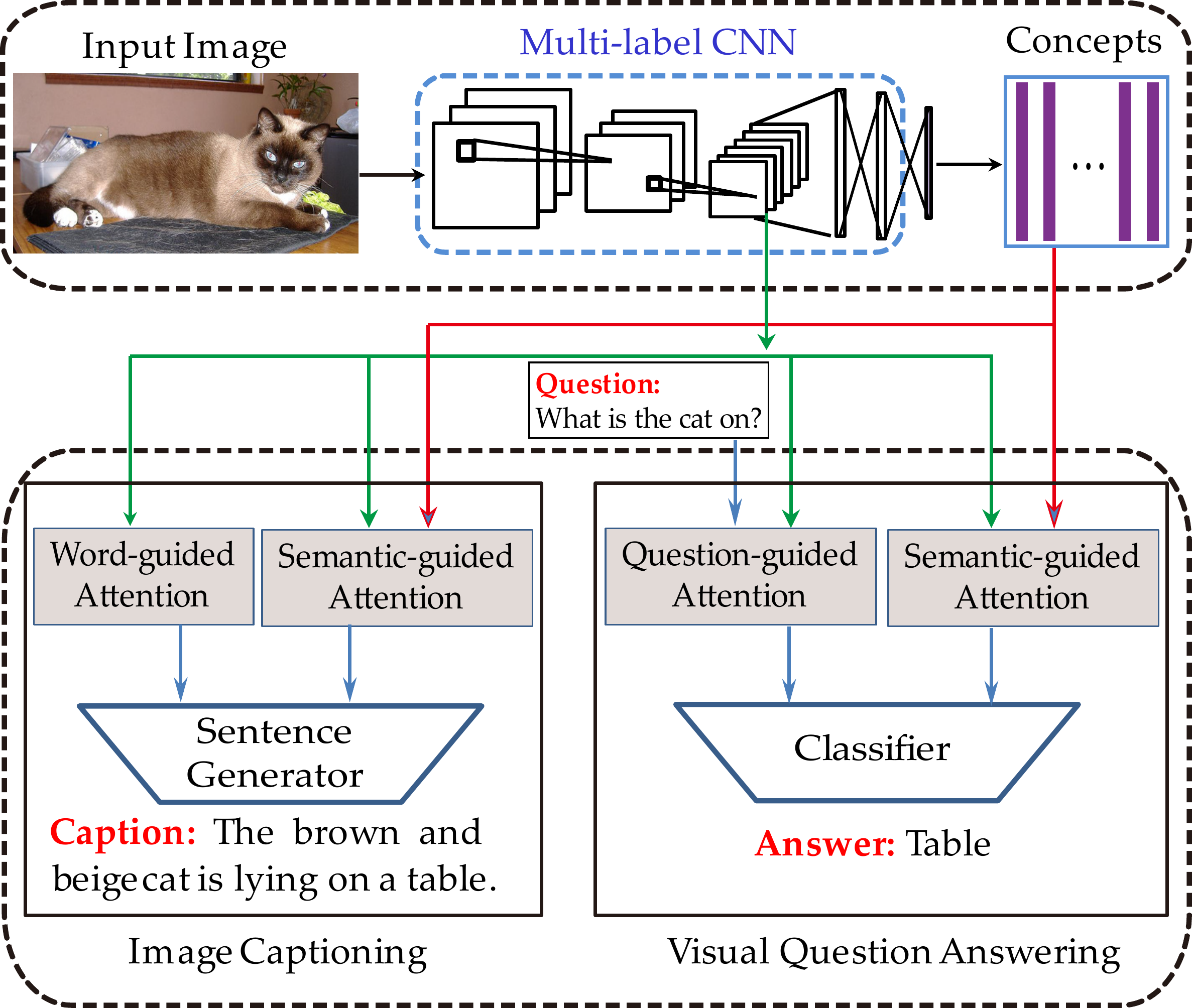}\\
  \caption{Overview of our scheme for vision-to-language tasks. The proposed approach is composed of two level attention modules. The first attention module is text-guided attention (\emph{i.e.}, word-guided attention for image captioning and question-guided attention for VQA) which is used to dig up the mapping relation between word and image region. Semantic-guided attention is the second attention module, which is used for explore the relationship between image and semantic concept. At last, the two parts of representations which output from the two attention modules are jointly embedded into multi-modal space to generate the corresponding captions or answers.}
  \label{introduction}
\end{figure}

Image high-level semantic concepts (also called image attributes, \textsl{i.e.}, objects, actions, scenes and object's attributes of images) are very important information for V2L tasks.
In previous works,  the spatial attention based methods (to distinguish the semantic-guided attention network, the spatial attention network is called text-guided attention network in this paper) are the most popular scheme for V2L tasks. Namely, the relationships between natural language elements with image regions via computing the attention weights between words/questions and image regions. The core idea of the spatial attention mechanism is that every word in captions or every question should only correspond to one or several regions of an image. Although the spatial attention based methods can dig up the subtle relationships between image and text elements,  high-level semantic concepts have not been fully utilized for V2L tasks in most of the previous works, while these concepts are important for humans when understanding a scene \cite{MLAN, DBLP:conf/cvpr/WuSLDH16}. Actually, the semantic concepts bridge images and natural language information together and can contribute significantly to eliminate the well-known semantic gap. That is because the semantic concepts are not only the important high-level visual information for images, but also the important component of captions.
For example, image in Fig. \ref{introduction} shows an brown cat. According to the semantic concepts in this paper, the words ``\emph{brown}'' and ``\emph{cat}'' are high-level information of the image, and they provide very important information for understanding the image. Meanwhile, the two words also the components of the corresponding captions. To eliminate the semantic gap between images and natural language, the high-level semantic information is an extra input of the proposed method.
\par Moreover, every high-level semantic concept should correlated to a specific image region. Some previous methods, such as \cite{DBLP:conf/cvpr/WuSLDH16, TPAMI/Q}, attempt to use the image semantic information to complete the V2L tasks, the semantic information is encoded into vectors and directly input into the language generating model. However, this process is not the optimal one because it cannot dig up the relationships between the semantic concepts and image regions. In this paper, a semantic-guided attention network is designed to explore the relationships between image semantic concepts and image regions. Namely, the image semantic concepts information is used to attend the corresponding regions.  Other works like \cite{MLAN} use the attention mechanism for image semantic information, but the guidance information is natural language. In other words, the high-level semantic information is attended the by the corresponding caption or the question. Actually, every semantic concept is more correlated to a specific image regions. For instance, image in Fig. \ref{introduction} shows a brown cat. Both the concept words ``\emph{cat}'' and ``\emph{brown}'' are important elements of the captions, while they correspond to the ``cat'' regions. So, exploring the relationships between image regions and image semantic concepts are more effective than the relationships between image semantic concepts and natural language elements.
\par Motivated by the aforementioned two reasons, we propose a methods with a semantic-guided attention network. The semantic-guided attention network contains two sub-parts which are used to highlight the concept-related regions and the region-related concepts, respectively. In addition, text-guided attention network is also reserved to explore the subtle relationships between image regions and natural language parts. For example, when describing the content of the image in Fig. \ref{introduction}, the phrase ``\emph{brown cat}'' should map with the ``\emph{cat}'' region of the image. For VQA task, the question is ``\emph{What is the cat on?}''. When answering this question, the ``\emph{cat}'' and its surrounding regions should be focused on because these regions are most related to the question.
So, to simultaneously learn the relationships between high-level semantic concepts and image regions and the correlations between natural language elements and image regions, we unify two sub-attention networks (semantic-guided attention network and tex-guided attention network) into a framework.
\par Fig. \ref{introduction} shows the overall scheme of the proposed approach. The approach mainly includes two level attention networks. One is the text-guided attention network which is used to select text-related regions. The text-guided attention network has two variants  for image captioning and VQA, respectively. In image captioning task, the text-guided attention network is called word-guided attention which is used to explore the relationships between words and image regions. In VQA task,  the text-guided attention network is called question-guided attention which is used to select the image regions corresponding to the question. The other is the semantic-guided attention network which is used to dig up the relationship between image regions and high-level semantic concepts. The outputs of these two networks are projected into the same multi-modal space to generate captions or answers.

\subsection{Contributions}
The core contributions can be summarized as follows:
\par 1) An approach based on image high-level semantic attributes and local image features is proposed to address the challenges of V2L tasks. Specially, the high-level semantic attributes information is used to reduce the semantic gap between images and text.
\par 2) An novel semantic-guided attention network is designed to explore the mapping relationships between semantic attributes and image regions. The semantic-guided attention network highlights the concept-related regions and selects the region-related concepts.
\par 3) Two special V2L tasks (\emph{i.e.}, image captioning and visual question answering) are addressed by the proposed approach. Taking into account their characteristics, two sub-models was designed for image captioning and VQA, respectively. Experimental results show that our models are effective for V2L tasks.

\subsection{Organization}
\par The rest of this paper is organized as follows. In Section \ref{Related Work}, some previous works are briefly introduced. Section \ref{our model} presents our approach for V2L tasks. To validate the proposed method, the experimental results are shown in Section \ref{Experiments}. At last, Section \ref{Conclusion} makes a brief conclusion for this paper.
\section{Related Work} \label{Related Work}
With the development of deep learning, some related and recent work on deep learning has been researched for visual content analysis \cite{DBLP:journals/tcsv/HanCLZ18, DBLP:journals/tcyb/HanCLYL18, DBLP:journals/tip/ChengHZX19}. In this section, some typical methods for V2L tasks, \emph{i.e.}, image captioning and VQA, are introduced.
\subsection{Image caption generation}
Using a natural language sentence to describe the content of the given image has long been researched in artificial intelligence. A traditional approach is to use predefined visual templates to generate sentences by filling detected visual concepts. Kuznetsova \textsl{et al.} \cite{DBLP:journals/pami/KulkarniPODLCBB13} pose the image caption generation task as a retrieval problem. They first retrieval a similar image and the corresponding descriptions from the training set, and compose a new sentence based on the retrieval descriptions.
Sentences generated by these methods are less variety and very limited, which cannot describe the contents of the test image very well.
\par Recent works using the deep neural networks has gained many encouraging results on image caption generating task. Mao \textsl{et al.} \cite{DBLP:journals/corr/MaoXYWY14a} proposed a \textsl{multi-modal recurrent neural network} (m-RNN) to explore the relationships between vision and text information. This model predicts the next word by computing the probability distribution of the next word conditioned on the previous words and visual features at each time-step.
Karpathy \textsl{et al.} \cite{DBLP:conf/cvpr/KarpathyL15} also proposed a multi-modal RNN model to generate sentences to describe the content of a given image. But in contrast to m-RNN, the image features are input into the multi-modal RNN only at the first time-step. Vinyals \textsl{et al.} \cite{DBLP:conf/cvpr/VinyalsTBE15} proposed a similar method, which combined deep CNN for image feature extracting with an LSTM for sentence generating. Donahue \textsl{et al.} \cite{DBLP:conf/cvpr/DonahueHGRVDS15} proposed an unified model for activity recognition, image captioning and video description. To generate captions for image, this model use multiple layers of LSTM. Wu \textsl{et al.} \cite{DBLP:conf/cvpr/WuSLDH16, TPAMI/Q} proposed a caption generation model based on attributes. They use the most common words as the semantic attributes. At the sentence generating step, not only the global image feature is input into RNN, but the semantic attribute vector also be used as one input of RNN.
\par Attention-based model becomes a hot topic on image caption generation. Xu \textsl{et al.} \cite{DBLP:conf/icml/XuBKCCSZB15} proposed an attention model to solve the image caption generation problem. In contrast to the previous models, it uses the output of last convolutional layer as the image features. Through flattening the feature map into 196 vectors, each vector denotes one region of the image. At each time-step, only one or several regions are selected by the attention mechanism.  \cite{DBLP:conf/cvpr/YouJWFL16} proposed an image captioning model with semantic attention. It uses a set of attribute detectors to get some semantic concepts and the attention mechanism can select specific items form these concepts. Fu \textsl{et al.} \cite{Kun_Fu} proposed a model based on spatial attention and scene-specific contexts.
\subsection{Visual question answering}
Malinowski \textsl{et al.} \cite{DBLP:conf/nips/MalinowskiF14} may be the first researchers to study the ``open-world'' visual question answering problem. They proposed a method with two important parts. One for semantic text parsing and the other is image segmentation with a Bayesian formulation to sample from nearest neighbors in the training set. This approach is very dependent on the human defined predicates and the accuracy of the image segmentation. Tu \textsl{et al.} \cite{DBLP:journals/ieeemm/TuMLCZ14} proposed a question answering based on joint parse graph from text and videos. All these early approaches have a common shortage: the answer is limited on the form of question.
\par Recently, deep neural network models have gained many encouraging results in the field of computer vision and natural language processing. Inspired by these encouraging results, an architecture based on ``CNN-RNN'' has become the most popular trend. Gao \textsl{et al.} \cite{DBLP:conf/nips/GaoMZHWX15} used CNN to encode the image. Another two RNNs are used to encode the question and generate the answer, respectively. Similar to \cite{DBLP:conf/nips/GaoMZHWX15}, Malinowski \textsl{et al.} also proposed a method based on ``CNN-RNN'' architecture. However, \cite{DBLP:conf/nips/GaoMZHWX15} only used one RNN as question encoder and decoding the image and question into answer. In \cite{DBLP:conf/nips/RenKZ15}, Ren \textsl{et al.} took the visual question answering as classification problem. Their method used the LSTM as the question encoder and the image was treated as the first world. The answer was generated from an classifier which is a softmax layer. The input of the softmax layer was the output of the last time-step of the LSTM. Wu \textsl{et al.} \cite{DBLP:conf/cvpr/WuSLDH16} proposed a method which contains two different LSTMs to encode the question together with decode it and image information into answer with multiple words. It is worth noting that this model used the global image feature and image attribute vector output from the attribute detector as image information. Their team also did another work. This work was more complicate than ever before because on the basis of \cite{DBLP:conf/cvpr/WuSLDH16}, they added external knowledge and caption vector as another two inputs to the encoder LSTM. They encoded five descriptions into vectors and pooled these vectors into one vector as caption vector. Noh \textsl{et al.} \cite{DBLP:conf/cvpr/NohSH16} used CNN with dynamic parameter prediction to solve the image question answering problem. To reduce the complexity of the problem, they incorporated a hashing technique to select the weights. \cite{DBLP:conf/aaai/MaLL16} proposed a model with CNN architectures for learning not only for image and question, but also their inter-modal relationships to produce the answering.
\par A limitation of the most aforementioned methods is that they only use global image feature to represent the input image. This may lead to some irrelevant or noisy information input into the answering module. To address the aforementioned problem, attention mechanism is widely used in question answering system.
A typical model is SANs \cite{DBLP:conf/cvpr/YangHGDS16} which is short for stacked attention networks. This model used semantic representation of a question to search for the corresponding regions in an image which related to the question and the answer. It also stacked the attention network because the authors argued that visual question answering needs to multiple steps of reasoning. Shih \textsl{et al.} \cite{DBLP:conf/cvpr/ShihSH16} presented a method which learns to answer questions by selecting image regions relevant to the questions. Unlike to \cite{DBLP:conf/cvpr/YangHGDS16}, which used one layer neural network to compute the attention distribution, this model mapped question queries and image features from various regions into a shared space through an inner product manuscription.
Xu \textsl{et al.} \cite{DBLP:conf/eccv/XuS16} proposed a spatial memory network to the visual question answering task. Their memory networks were recurrent neural networks with attention mechanism that choose relevant regions stored in memory.
\cite{DBLP:conf/cvpr/NamHK17} presented an dual attention model which jointly used visual and textual attention to capture the fine-grained relationship between vision and language.
Lu \textsl{et al.} \cite{DBLP:conf/nips/LuYBP16} proposed a co-attention for both image and question. Different from the most above models, this model used an hierarchical question encoding.
Kazemi \textsl{et al.} \cite{DBLP:journals/corr/KazemiE17} proposed a strong baseline for visual question answering. Their model used two-layer convolutional neural network to realize the stacking attention and produce probabilities over answer classes.
Yu \textsl{et al.} \cite{MLAN} presented a multi-level attention model which contained context-aware visual attention and semantic attention modules. The context-aware module used a question to select relevant regions and the semantic attention module aimed to find important concepts.
Xiong \textsl{et al.} \cite{DBLP:conf/icml/XiongMS16} proposed a model named dynamic memory network which mainly contained two important parts: input module and episodic memory module. The core component of the input module is the bidirectional gated recurrent unit which was used to explore the relationship between local regional image features. In fact, the episodic memory module is also an attention module, which extracted a contextual vector based upon the current focus.


\begin{figure*}[ht]
  \centering
  \includegraphics[width=0.90\linewidth]{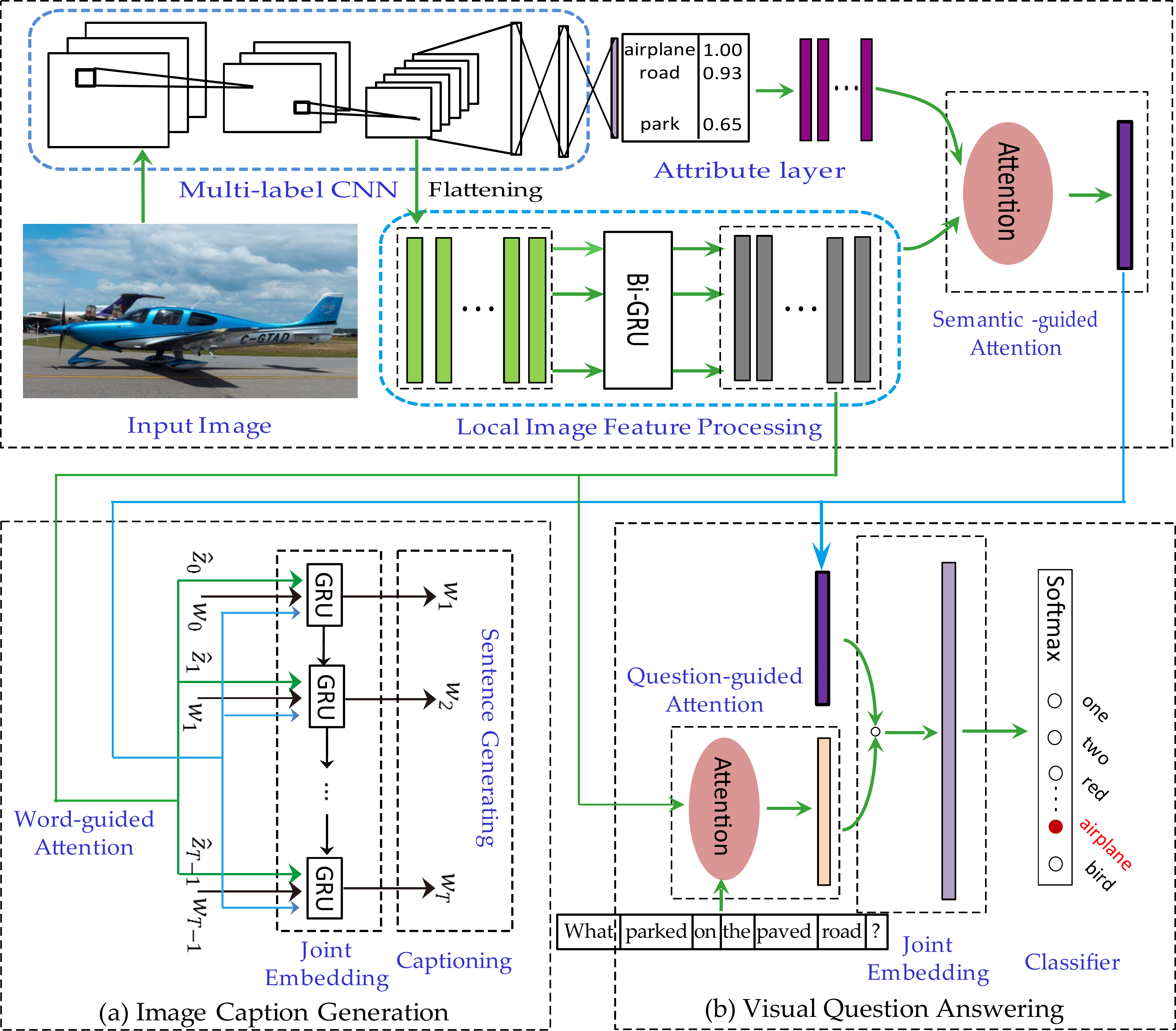}\\
  \caption{Overall framework of our approach for two typical Vision-to-Language tasks: (a) image captioning and (b) visual question answering. Both the two sub-frameworks have 1) a multi-label CNN, 2) an attribute layer, 3) a bidirectional GRU module, 4) semantic-guided attention network, 5) text-guided attention network and 6) joint embedding layer. The multi-label CNN is used as local image feature extractor as well as the attribute classifier. The bidirectional GRU module is used to explore the relationships between the local image features. The semantic-guided attention network is designed to highlight the concept-related regions and select the region-related concepts. The text-guided attention network explores the fine-grained mapping relationship between language element and image region. All the information is fused in the joint embedding layer. At last, the multi-modal information is used to generate the caption or the answer.}
  \label{V2L_fig}
\end{figure*}

\section{Proposed Approach} \label{our model}

 Fig. \ref{V2L_fig} shows the overall framework of our approach for image captioning and VQA, two typical V2L tasks. Both the two sub-frameworks consist of six part: 1) a multi-label CNN, 2) an attribute layer, 3) a bidirectional GRU module, 4) semantic-guided attention network, 5) text-guided attention network and 6) joint embedding layer. The first two modules are used to extract image attributes and local features. The bidirectional GRU \cite{DBLP:conf/emnlp/ChoMGBBSB14} module is used to explore the relationships among the local image features. The concept vectors output from the attribute layer and the proposed local image feature output from the bidirectional GRU are input into semantic-guided attention network. This is designed to highlight the concept-related regions and select the region-related concepts. The text-guided attention network explores the fine-grained mapping relationship between language elements and image regions. All the information is fused in the joint embedding layer. At last, the multi-modal information is used to generate the caption or the answer. There are two little differences between the two sub-frameworks. First, image captioning is considered as a generating problem, which means the captions are generated word by word. While VQA is treated as a classification problem, which all the answers are processed as class labels. Second, the text-guided attention network in image captioning and VQA are called word-guided attention network and question-guided attention network, respectively.

\subsection{Image Concepts Predicting}\label{image_concept}
To train an image concepts predictor, the concepts vocabulary should be built at first. Similar to \cite{MLAN}, we collect all words from the MS COCO image captioning dataset \cite{lin2014coco}. All words are reverted to the prototype (\emph{i.e.}, the form of nouns and the tense of verbs are not sensitive). To select the concept words, the word frequencies are counted at first. And then the meaningless words (\emph{e.g.}, ``\emph{a}'', ``\emph{is}'', ``\emph{on}'' and so on) are abandoned. After the rough screening, we select the $c$ most frequent words as the image semantic concepts candidate.

\par After constructing the concept vocabulary, we label each image with a $c$-dimensional vector through comparing the captions with the concept vocabulary. We then train the concept detector (\emph{i.e.}, Multi-label CNN in Fig. \ref{V2L_fig}). As a results, each image $I$ is represented as a concept vector ${{v}_I} = {\left[ {\begin{array}{*{20}{c}}
{{v_{I1}}}&{{v_{I2}}}& \cdots &{{v_{Ic}}}
\end{array}} \right]^T} \in {\mathbb{R}^{c}}$
and each element denotes the probability of the corresponding concept. To train the concept predictor, the last softmax layer of the single label CNN is replaced by the sigmoid cross entropy loss layer. Suppose that there are $N$ training samples and ${y_i} = {\left[ {\begin{array}{*{20}{c}}
{{y_{i1}}}&{{y_{i2}}}& \cdots &{{y_{ic}}}
\end{array}} \right]^T}$ (where $y_{ij}=0\ or\ 1, i=1,2,\cdots,N, j=1,2,\cdots,c$) is the attribute label of $i$-th image. And the loss function is defined as follows:
\begin{equation}
{L_M} =  - \frac{1}{N}\sum\limits_{i = 1}^N {\sum\limits_{j = 1}^c {\left[ {{y_{ij}}\log v_{Ij}^{(i)} + \left( {1 - {y_{ij}}} \right)\log \left( {1 - v_{Ij}^{(i)}} \right)} \right]} } .
\end{equation}
\par According to $v_I$, the concept set $v_c$ of image $I$ is defined as follows:
\begin{equation}\label{v_c}
{{v}_c} = \left\{ {{v_{ci}}} \right\},{v_{ci}} = {e_i} \cdot \delta \left( {{v_{Ii}} \ge \varepsilon } \right),i \in \left\{ {1,2, \cdots ,c} \right\},
\end{equation}
where $\delta \left(  \cdot  \right)$ denotes the indicator function, $\varepsilon$ is a threshold and we set it as 0.6 in this paper, ${e_i} \in {\mathbb{R}^c}$ is a vector where the $i$-th element equal to 1 and the other elements equal to 0.
\subsection{Local Image Feature Processing}\label{local_image_representation}
As illustrated in \cite{DBLP:conf/icml/XiongMS16}, we use a pre-trained CNN (\emph{i.e.}, VGG-19 in this paper) to extract local image features. When a raw image $I$ is input into VGG-19, we flat the feature map output from the CONV5-4 layer. The process can be written as follows:
\begin{equation}\label{v_l}
{{v}_l} = \left\{ {{v_{l1}},{v_{l2}},\cdots,{v_{lC}}} \right\} = flatten\left( {Conv(I)} \right),
\end{equation}
where ${v}_{li}\in \mathbb{R}^{D},\ i\in \{0,1,\cdots ,C\}$ denotes the feature of $i$-th location of image $I$. In other words, each image $I$ is divided into $C$ locations and every ${v}_{li}$ represents one location. So, we call ${v}_{li}$ is the location feature representation.

\par
\begin{figure}[!t]
  \centering
  \includegraphics[width=0.8\linewidth]{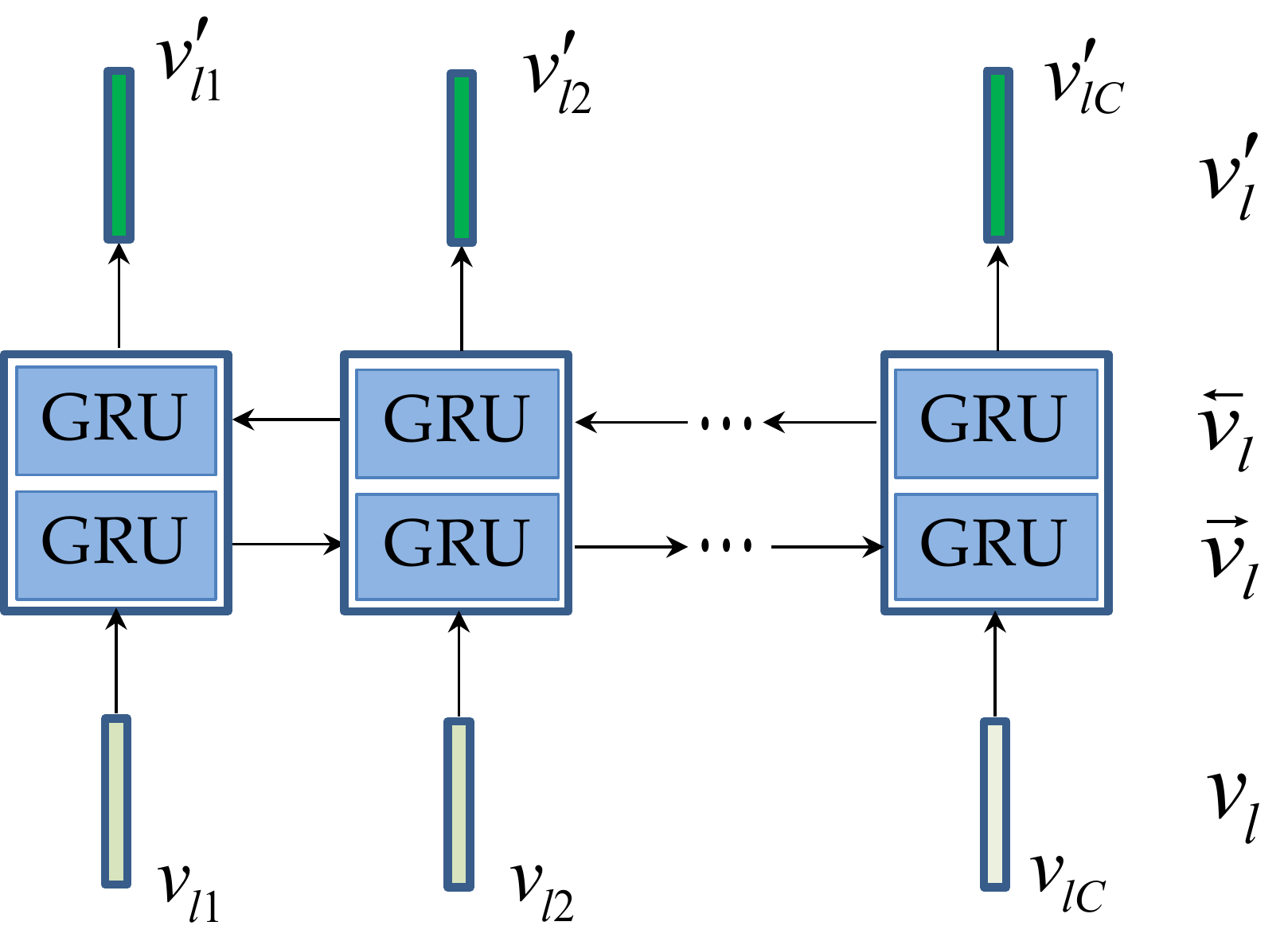}\\
  \caption{An illustration of local image processing. The local image feature output from convolutional layer are encoded into bi-directional GRU. The sum of hidden states of the forward and backward GRU is denoted as the context-aware visual representation.}
  \label{local_image_feature}
\end{figure}

 The local image feature extracted from above do not yet have global information available for them. Without global information, their representational power is quite limited because it suffers from the simple issues like locational variance causing accuracy problems or object scaling. According to \cite{DBLP:conf/icml/XiongMS16}, the bidirectional RNN can solve the aforementioned problem. Following this idea, we use a bidirectional GRU to explore the relationship among regions (As illustrated in Fig. \ref{local_image_feature}). The formulas are shown as follows:
\begin{equation}\label{v_l'}
\begin{array}{l}
{{\vec v}_{li}} = GR{U_f}\left( {{v_{li}},{{\vec v}_{li - 1}}} \right)\\
{{\mathord{\buildrel{\lower3pt\hbox{$\scriptscriptstyle\leftarrow$}}
\over v} }_{li}} = GR{U_b}( {{v_{li}},{{\mathord{\buildrel{\lower3pt\hbox{$\scriptscriptstyle\leftarrow$}}
\over v} }_{li + 1}}} )\\
{v'_{li}} = {{\vec v}_{li}} + {{\mathord{\buildrel{\lower3pt\hbox{$\scriptscriptstyle\leftarrow$}}
\over v} }_{li}}
\end{array},
\end{equation}
where ${\vec v}_{li}$ and ${\mathord{\buildrel{\lower3pt\hbox{$\scriptscriptstyle\leftarrow$}} \over v} }_{li}$ are the hidden states of forward and backward GRU at time-step $i$, respectively. At last, the sum of ${\vec v}_{li}$ and ${\mathord{\buildrel{\lower3pt\hbox{$\scriptscriptstyle\leftarrow$}} \over v} }_{li}$ denotes the context-aware visual representation of the $i$-th image region ${v'_{li}}$. We denote ${v'_l} = \left[ {\begin{array}{*{20}{c}}
{{v'_{l1}}}&{{v'_{l2}}}& \cdots &{{v'_{lC}}}
\end{array}} \right]$.
\subsection{Semantic-guided Attention}\label{SGA}
To find the fine-grained relationship between image regions and semantic concepts, we propose a semantic-attention network. We connect the visual representation $v'_{l}$ and semantic concepts representation ${{v}_c}$ by similarity between them at all image-concepts and concept-regions. Specifically, given an image representation $v'_{l}\in \mathbb{R}^{D \times C}$, and the concepts representation ${{v}_c}\in \mathbb{R}^{c \times c}$, the similarity matrices are calculated as follows:
\begin{equation}\label{sim}
\begin{array}{l}
P = {\left( {{W_l}{v'_l} \oplus {{b}_l}} \right)^T} \cdot \left( {{W_c}{{v}_c} \oplus {{b}_c}} \right)\\
P' = {\left( {{W'_c}{{v}_c} \oplus {b'_c}} \right)^T} \cdot \left( {{W'_l}{v'_l} \oplus {b'_l}} \right)
\end{array},
\end{equation}
where $P\in \mathbb{R}^{C \times c}$ and $P'\in \mathbb{R}^{c \times C}$ represent the similarity matrices between image regions and concepts. Concretely, ${P_{ij}}$ and ${P'_{i'j}}$ are scores which represent the similarity of the $j$-th concept with $i$-th region representation and the similarity of the $j$-th region representation with $i'$-th concept, respectively. ${W_l} \in {\mathbb{R}^{d \times D}}$, ${W_c} \in {\mathbb{R}^{d \times c}}$, ${W'_c} \in {\mathbb{R}^{d' \times c}}$, ${W'_l} \in {\mathbb{R}^{d' \times D}}$, $b_l \in {\mathbb{R}^d }$, $b_c \in {\mathbb{R}^d }$, $b'_c \in {\mathbb{R}^{d'} }$, and $b'_l \in {\mathbb{R}^{d'} }$ denote weights parameters. Note that ``$\oplus$'' represents the addition of a matrix and a vector. The addition between a matrix and a vector is performed by adding each column of the matrix by the vector.

\par After calculating the similarity matrices, the formula of attention weights is as follows:
\begin{equation}\label{alp}
\begin{array}{l}
\alpha _i^l = \frac{{\exp \left( {{{\max }_j}\left( {{P_{ij}}} \right)} \right)}}{{\exp \left( {\sum\nolimits_k {{P_{ik}}} } \right)}},\left( {i \in \{ 1,2, \cdots ,C\} } \right)\\
\alpha _{i'}^c = \frac{{\exp \left( {{{\max }_j}\left( {{P'_{i'j}}} \right)} \right)}}{{\exp \left( {\sum\nolimits_k {{P'_{i'k}}} } \right)}},\left( {i' \in \{ 1,2, \cdots ,c\} } \right)
\end{array}.
\end{equation}
\par Based on the above attention weights, the concept-based image region representation and the region-based concept representation are calculated as follows:
\begin{equation}\label{v}
\begin{array}{l}
{{\hat v}_l} = \sum\limits_{i = 1}^C {\alpha _i^l{{v'}_{li}}} ,\\
{{\hat v}_c} = \sum\limits_{i' = 1}^c {\alpha _{i'}^c{v_{ci'}}} .
\end{array}
\end{equation}
\par After computing the weighted image and concept representations, we concatenate them into a vector which contains the image feature and semantic concepts representation. The formula is:
\begin{equation} \label{v_II}
{v'_I} = \left[ {{{\hat v}_l};{{\hat v}_c}} \right].
\end{equation}
\subsection{Image Captioning} \label{IC}
The proposed model for image captioning is summarized in Fig. \ref{V2L_fig} (a). Similar to \cite{DBLP:conf/cvpr/VinyalsTBE15}, our language generation model is trained by maximizing the probability of the correct description conditioned on the given image. Combined with our model, the log-likelihood function can be written as follows:
\begin{equation} \label{log}
\log P\left( {S\left| {{{v}_l}, {v'_I}} \right.} \right) = \sum\limits_{t = 1}^T {\log P\left( {{w_t}\left| {{w_{1:t - 1}},{{v}_l},{v'_I}} \right.} \right)} ,
\end{equation}
where $S = \left\{ {{w_0},{w_1}, \cdots ,{w_T}} \right\}$ is the description of image $I$, $w_t$ is the $t$-th word of the sentence $S$, and $T$ is the length of the sentence. Based on Eq. (\ref{log}), the probability of generating the word $w_t$ (\emph{i.e.}, $P\left( {{w_t}\left| {{w_{1:t - 1}},{v'_I}} \right.} \right)$) is determined by the output of the semantic-guided attention network ${v'_I}$ and the previous words ${w_{0:t - 1}}$. We exploit GRU to model this.
\subsubsection{Sentence Representation} \label{Sentence Representation}
In our model, we encode words into one-hot vectors. For example, the benchmark dataset has $N_{0}$ different words, and every word is encoded into a $N_{0}$-dimension vector in which only one value equals to 1 and others equal to 0. When a raw image is input into our model, a corresponding sentence $S$ is generated which is encoded as a sequence of one-hot vectors. We denote $S=({w_1}, {w_2},\cdots ,{w_T})$, where ${w}_i\in \mathbb{R}^{N_0}$ represents the $i$-th word in the sentence. We project these words into embedding space. The concrete formula is as follows:
\begin{equation}\label{sentence representation}
  {{\bf{s}}_t} = {W_s}\cdot {w_t},\ t\in \{1,2,\cdots ,N\},
\end{equation}
where $W_s$ is the embedding matrix of sentences which projects the word vector into the embedding space. So the projection matrix $W_s$ is a $N_0\times h$ matrix where $N_0$ is the size of the dictionary and $h$ is the dimensionality of the embedding space.
\subsubsection{Word-guided Attention} \label{Word-guided Attention}
Similar to \cite{DBLP:conf/icml/XuBKCCSZB15}, we use word-guided attention mechanism for local feature. At each time-step, the attention mechanism uses the previous hidden state $h_{t - 1}$ which concludes the previous words information to decide the local feature. The attention model is defined as follows:
\begin{equation}\label{att}
\begin{array}{l}
{{{\bf{\tilde \alpha }}}_t} = \tanh \left[ {{{\left( {{\bf{w}}_a^T{v'_l}} \right)}^T} + {U_a}h_{t - 1} + {b_a}} \right]\\
{{\bf{\alpha }}_t} = {\rm{softmax}}\left( {{{{\bf{\tilde \alpha }}}_t}} \right) \buildrel \Delta \over = {\left[ {\begin{array}{*{20}{c}}
{{\alpha _{t1}}}& \cdots &{{\alpha _{tC}}}
\end{array}} \right]^T}
\end{array},
\end{equation}
where ${{\bf{w}}_a}\in {\mathbb{R}^{D}}$ and ${U_a}\in {\mathbb{R}^{C\times h}}$ are weights, ${b_a}\in {\mathbb{R}^{C}}$ is bias. ${{\alpha }_t}\in{\mathbb{R}^C}$ is a probability vector whose each value denotes the probability of the corresponding local image feature. In our algorithm, we use the soft attention model. Therefore, ${{\bf{\hat z}}_t}$, the word-related region representation at time-step $t$, is calculated as follows:
\begin{equation}\label{lv}
{{\hat z}_t} = {v'_l}{{\bf{\alpha }}_t} = \sum\limits_{i = 1}^C {{\alpha _{ti}}{v'_{li}}} .
\end{equation}
\par Through Eq. (\ref{lv}) we know that ${{\bf{\alpha }}_t}$ decides which locals should be emphasized at the current time-step.
\subsubsection{Gate for $v'_I$} \label{Gate for ${v'}_I$}
To control when and how much $v'_I$ should be input into sentence generation GRU, we design a gate to achieve it. The gate is defined as follow:
 \begin{equation}\label{gate-global}
 {g_t} = \sigma ({\bf{w}}_g^T{h}_{t - 1} + {b_g}),
 \end{equation}
 where ${\bf{w}}_g \in {\mathbb{R}^{h}}$ is weight vector and $b_g \in {\mathbb{R}}$ is bias.
 After calculating the gate, the $v'_I$ is controlled  as following formula:
  \begin{equation}\label{gg}
 {v'_t} = {g_t}{v'_I}.
 \end{equation}
 \subsubsection{Sentence Generating}
 After getting ${{\hat z}_t}$ and $v'_t$, we use GRU to generate description for the given image. The formulae are as follows:
 \begin{equation}
{h_t} = GRU\left[ {{{\hat z}_t},{v'_t},{h_{t - 1}}} \right],
 \end{equation}
  \begin{equation} \label{p}
{p_{t + 1}}{\rm{ }} = {\rm{softmax}} ({h_t}).
 \end{equation}
 The loss function is written as follows:
 \begin{equation}
 \begin{array}{c}
L_C =  - \frac{1}{N}\sum\limits_{i = 1}^N {\log P\left( {{S^{(i)}}\left| {v_l^{(i)},{v'_I}^{(i)}} \right.} \right)} \\
 =  - \frac{1}{N}\sum\limits_{i = 1}^N {\sum\limits_{t = 1}^{{T^{(i)}}} {\log {p_t}\left( {w_t^{(i)}} \right)} }
\end{array},
 \end{equation}
 where $N$ is the number of training images and $T^{(i)}$ is the length of the sentence for the $i$-th training image. ${p_t}\left( {w_t^{(i)}} \right)$ equals to $p_{t+1}$ in Eq. (\ref{p}).
\subsection{Visual Question Answering} \label{VQA}
The model for VQA is illustrated in Fig. \ref{V2L_fig} (b). Similar to \cite{DBLP:conf/nips/RenKZ15}, we take VQA as a classification task. So, all the information should be jointly embedded into a classifier. Given the image $I$ and corresponding question $Q$, we expect the probability of the correct answer to reach maximum. The object function can be written as follows:
\begin{equation}
\hat A = \mathop {\arg \max }\limits_A P\left( {A\left| {{v_q},{v'_I}} \right.} \right),
\end{equation}
where $v_q \in \mathbb{R}^{q}$ denotes the representation of question $Q$, $v'_I$ represents the image feature and attribute representation of image $I$.
After encoding each question into a vector $v_q$, we calculate the question-guided region representation.
\subsubsection{Question-guided Attention Network}
Similar to word-guided attention in Section \ref{Word-guided Attention}, we design a question-guided attention network to select the question-related regions which can improve the accuracy of the answer. The formula of the attention weight is as follows:
\begin{equation}
{\alpha _i} = \frac{{\exp \left( {\sigma \left( {\left\langle {{W_q}{v_q},{W_l}{v'_{li}}} \right\rangle } \right)} \right)}}{{\sum\nolimits_k {\exp \left( {\sigma \left( {\left\langle {{W_q}{v_q},{W_l}{v'_{lk}}} \right\rangle } \right)} \right)} }},
\end{equation}
where $\left\langle { \cdot , \cdot } \right\rangle $ is the inner product operation symbol, ${W_q} \in {\mathbb{R}^{h' \times q}}$ and ${W_l} \in {\mathbb{R}^{h' \times D}}$ are projection matrices which project the question and region representation into the $h'$-dimensional multi-modal space. After that, the question-related region can be represented as follow:
\begin{equation} \label{v_lq}
{v_{lq}} = \sum\limits_{i = 1}^C {{\alpha _i}{v'_{li}}} ,{v'_{lq}} = \left[ {{v_q};{v_{lq}}} \right].
\end{equation}
\subsubsection{Joint Embedding}
Finally, we feed all the vectors (\emph{i.e.}, $v'_I$ and $v'_{lq}$) into classifier with an joint embedding layer to generate the answer. This can be represented as the following formulae:
\begin{equation}
\begin{array}{l}
u = \tanh \left( {W{v'_I} + U{v_{lq}} + b} \right)\\
{P_a} = {\rm{softmax}}(u)
\end{array},
\end{equation}
where $W$, $U$ and $b$ are the parameters of the last parameter layer, the input of the classifier, $v'_I$ and $v_{lq}$ are calculated in Eq. (\ref{v_II}) and Eq. (\ref{v_lq}), respectively, $P_a$ is the distribution of probability of answer candidates. The answer is the maximum probability of the candidates.
\par The loss function can be written as follows:
\begin{equation}
L_A =  - \frac{1}{N}\sum\limits_{i = 1}^N {\log P_a^{(i)}} ,
\end{equation}
where $N$ is the number of the train examples.
\section{Experiment} \label{Experiments}
\subsection{Train Details and Experimental Setup}
This section mainly shows the training details and the parameter setting. For both image captioning and VQA tasks, the variants of our models are trained with stochastic gradient descent \cite{DBLP:conf/icml/Zhang04} within adaptive learning rates. Specially, for the Flickr30K  \cite{DBLP:journals/tacl/YoungLHH14}, MS COCO \cite{lin2014coco}, VQA \cite{DBLP:conf/iccv/AntolALMBZP15}  and COCO-QA \cite{DBLP:conf/nips/RenKZ15}, Adam algorithm is used. For Flickr8K \cite{DBLP:journals/jair/HodoshYH(A), DBLP:conf/ijcai/HodoshYH(B)}, RMSProp is used to train the models. The parameter setting is shown in the following subsections.
\subsubsection{Local Image Feature}
In the proposed model, deep features generated from the CONV5-4 layer of VGG-19 are used to represent the images. The dimensionality of the feature map output from the Conv5-4 layer is $14 \times 14 \times 512$. Through flattening operation, the feature map is transformed into $196 \times 512$. So, in Section \ref{local_image_representation}, the parameters $C = 196$ and $D = 512$.
\subsubsection{Image Concepts Encoding}
In the proposed method, $512$ concepts word are collected from the MS COCO image captioning dataset. So, each image is encoded into a $512$-dimensional vector. In other words, $c = 512$ in Section \ref{image_concept}.
\subsubsection{Word Encoding}
 In our model, we encode words into one-hot vectors. For example, the benchmark dataset has $M$ different words, every word is encoded into a $M$-dimension vector, in which only one value is equal to 1 and others are equal to 0. So the location of 1 in the vector denotes the corresponding word in the dictionary. It implies that $N_{0}$ in Section \ref{Sentence Representation} equals $M$. Specially, after filtering words less than 5 times in the training set, the value of $N_0$ equals 2538, 7414 and 8791 words for Flickr8K, Flickr30K and MS COCO, respectively.
\subsubsection{Question Encoding}
 To encode the questions, we first cast all question words which appear at least twice in the training and validation sets into lowercase. After collecting the question words vocabulary, each word is represented as one-hot vector. We use one layer Gated Recurrent Unit (GRU) with $512$-dimensionality hidden state to encode the question, and the last hidden state of the GRU as the question representation. So the parameter $q=512$ in Section \ref{VQA}.
\subsubsection{Other Parameters}
In this paper, to calculate the relationship between the concept word and image regions, the parameter $d$ and $d'$ are introduced (In Section \ref{SGA}). In the experiments, we set $d'=d=512$. For image caption, the hidden state's dimensionality of the captioning GRU is set as 512 (\textsl{i.e.} $h = 512$ in Section \ref{IC}).
\subsection{Image Captioning}
\subsubsection{Dataset and Evaluation Metrics}
\textbf{Dataset.} We report results on the most popular three datasets: Flickr8K, Flickr30K and MS COCO. Among them, Flickr8K and Flickr30K have 8,092 and 31,783 images respectively, and each image has 5 reference sentences. MS COCO dataset has 123,287 images and the most images has 5 reference sentences. Before the experiment, we preprocess the datasets as \cite{DBLP:conf/cvpr/KarpathyL15} did. First, we convert all letters of sentences to lowercase, remove non-alphanumeric characters and get rid of words that occur less than five times on the training set. Second, we discard these data which have more than 5 corresponding sentences to guarantee that every image has the same number of describing sentences. For MS COCO, we evaluate our model with the widely used publicly available splits in \cite{DBLP:conf/cvpr/KarpathyL15}.
\par \textbf{Evaluation Metrics.} We report results with the BLEU \cite{papineni2002bleu}, METEOR \cite{banerjee2005meteor} and CIDEr \cite{DBLP:conf/cvpr/VedantamZP15} metrics which are the most frequently used in the caption generation literature. The first two metrics are originally designed for evaluating the quality of the automatically machine translation. BLEU score represents the precision ratio of the generated sentence compared with the reference sentences. METEOR score reflects the precision and recall ratio of the generated sentence. It is based on the harmonic mean of uniform precision and recall. CIDEr measures consistency between n-gram occurrences in generated and reference sentences, where this consistency is weighted by n-gram saliency and rarity. For BLEU, we report the scores from BLEU-1 to BLEU-4, which denote the precision of N-gram (N equals to 1, 2, 3 and 4). For both metrics, the higher score they are, the higher quality of the generated sentences they have.
\subsubsection{Results on Flickr8K and Flickr30K}
\begin{table*}[t]
\renewcommand{\arraystretch}{1}
\caption{Experimental results on Flickr8K \& Flickr30K.}
\label{results_on _Flickr}
\begin{center}
\resizebox{\linewidth}{!}{
\newcommand{\tabincell}[2]{\begin{tabular}{@{}#1@{}}#2\end{tabular}}
\small\begin{tabular}{c||cccccc|cccccc}
  \thickhline
  &\multicolumn{6}{c|}{\textbf{Flickr8K}}  &\multicolumn{6}{c}{\textbf{Flickr30K}}\\
  \cline{2-13}\raisebox{0.5em}{Model} & \tabincell{c}{B-1} & \tabincell{c}{B-2} & \tabincell{c}{B-3} & \tabincell{c}{B-4} & \tabincell{c}{METEOR} & \tabincell{c}{CIDEr} & \tabincell{c}{B-1} & \tabincell{c}{B-2} & \tabincell{c}{B-3} & \tabincell{c}{B-4} & \tabincell{c}{METEOR} & \tabincell{c}{CIDEr}\\
  \hline\hline
  NeuralTalk \cite{DBLP:conf/cvpr/KarpathyL15}  & 57.9 & 38.3 & 24.5 & 16.0 & 16.7 & 31.8 & 57.3 & 36.9 & 24.0 & 15.7 & 15.3 & 24.7\\
  Google-NIC \cite{DBLP:conf/cvpr/VinyalsTBE15} & 63.0 & 41.0 & 27.0 & - & - & - & 66.3 & 42.3 & 27.7 & 18.3 & - & -\\
  m-RNN \cite{DBLP:journals/corr/MaoXYWY14a} & 56.5 & 38.6 & 25.6 & 17.0 & - & - & 60 & 41 & 28 & 19 & - & -\\
  \hline\hline
  Att-SVM + LSTM \cite{TPAMI/Q} & 73 & 53 & 38 & 26 & - & - & 68 & 49 & 33 & 23 & - & -\\
  Att-GlobalCNN + LSTM \cite{TPAMI/Q} & 72 & 53 & 38 & 27 & - & - & 70 & 50 & 35 & 27 & - & -\\
  \hline\hline
  NIC-VA-Soft \cite{DBLP:conf/icml/XuBKCCSZB15} & 67.0 & 44.8 & 29.9 & 19.5 & 18.9 & - & 66.7 & 43.3 & 28.8 & 19.1 & 18.5 & -\\
  NIC-VA-Hard \cite{DBLP:conf/icml/XuBKCCSZB15} & 67.0 & 45.7 & 31.4 & 21.3 & 20.3 & - & 66.9 & 43.9 & 29.6 & 19.9 & 18.5 & -\\
  ATT-$k$-NN \cite{DBLP:conf/cvpr/YouJWFL16} & - & - & - & - & - & - & 61.8 & 42.8 & 29.0 & 19.5 & 17.2 & -\\
  ATT-RK \cite{DBLP:conf/cvpr/YouJWFL16} & - & - & - & - & - & - & 61.7 & 42.4 & 28.6 & 19.3 & 17.7 & -\\
  ATT-FCN \cite{DBLP:conf/cvpr/YouJWFL16} & - & - & - & - & - & - & 64.7 & 46.0 & 32.4 & 23.0 & 18.9 & -\\
  RA \cite{Kun_Fu} & 59.5 & 40.4 & 26.2 & 16.6 & 17.8 & 39.9 & 62.9 & 44.1 & 30.6 & 21.0 & 18.7 & 43.2\\
  SS \cite{Kun_Fu} & 62.2 & 44.0 & 30.1 & 20.2 & 20.0 & 51.2 & 63.2 & 44.0 & 29.9 & 20.4 & 18.3 & 38.9\\
  RA + SS \cite{Kun_Fu} & 61.3 & 43.0 & 29.6 & 19.8 & 19.5 & 48.9 & 63.5 & 44.7 & 31.1 & 21.4 & 19.2 & 44.8\\
  \hline\hline
  Ours & 72.9 & 53.4 & 41.2 & 30.7 & 27.9 & 54.3 & 71.8 & 52.0 & 39.4 & 28.8 & 27.2 & 50.1\\
  \thickhline
\end{tabular}}
\end{center}
\end{table*}

We compare our method with several state-of-the-art methods on the Flickr8K and Flickr30K datasets. The contrast models can be roughly divided into three categories. The first category, such as NeralTalk \cite{DBLP:conf/cvpr/KarpathyL15}, Google-NIC \cite{DBLP:conf/cvpr/VinyalsTBE15} and m-RNN \cite{DBLP:journals/corr/MaoXYWY14a} in Table \ref{results_on _Flickr}, only uses the global image feature extracted by CNN, and only the feature is input into the sentence generator RNN. The second category is attribute-based models, which the global image feature and attribute vector are used to sentence generating. In Table \ref{results_on _Flickr}, Att-SVM + LSTM \cite{TPAMI/Q} and Att-GlobalCNN + LSTM \cite{TPAMI/Q} belong to this category. The third category is attention-based models. The attention-based model try to explore the relationship between image regions and words. NIC-VA \cite{DBLP:conf/icml/XuBKCCSZB15}, ATT \cite{DBLP:conf/cvpr/YouJWFL16} and RA \cite{Kun_Fu} \textsl{et al.} in Table \ref{results_on _Flickr} are all attention-based models.
Table \ref{results_on _Flickr} reports the image captioning results on the Flickr8K and Flickr30K. Between the contrast models, the attribute-based models show better performance than attention-based models. Specially for the Flikr8K, the Att-GlobalCNN + LSTM brings significant improvements nearly $5\%$ for B-1, $8\%$ for B-2, $8\%$ for B-3 and $7\%$ for B-4 on average. And the similarity improvements on the Flickr30K dataset. The phenomenon implies that the high-level semantic information (\textsl{i.e.}, attributes) is very important for image captioning task. Compared with the basic models (\textsl{i.e.}, none attribute and none attention are used), the attention-based models show much better performance for image captioning. The main reason is that the attention-based models can dig up the relationship between the image regions and sentence elements.
\par Although the state-of-the-art attribute-based models and attention-based models show good performance on the Flickr8K and Flickr30K datasets, Table \ref{results_on _Flickr} shows that our model gains a much better results on these datasets (only the B-1 score less than Att-SVM + LSTM on the Flickr8K dataset). The main reason is our model combines the semantic information and attention mechanism masterly. Specially, both the semantic information vector output from the semantic-guided attention network and the image region feature are selected by the word-guided attention network are used to generate the description sentence.
\subsubsection{Results on MS COCO}
\begin{figure*}[t]
  \centering
  \includegraphics[width=0.98\linewidth]{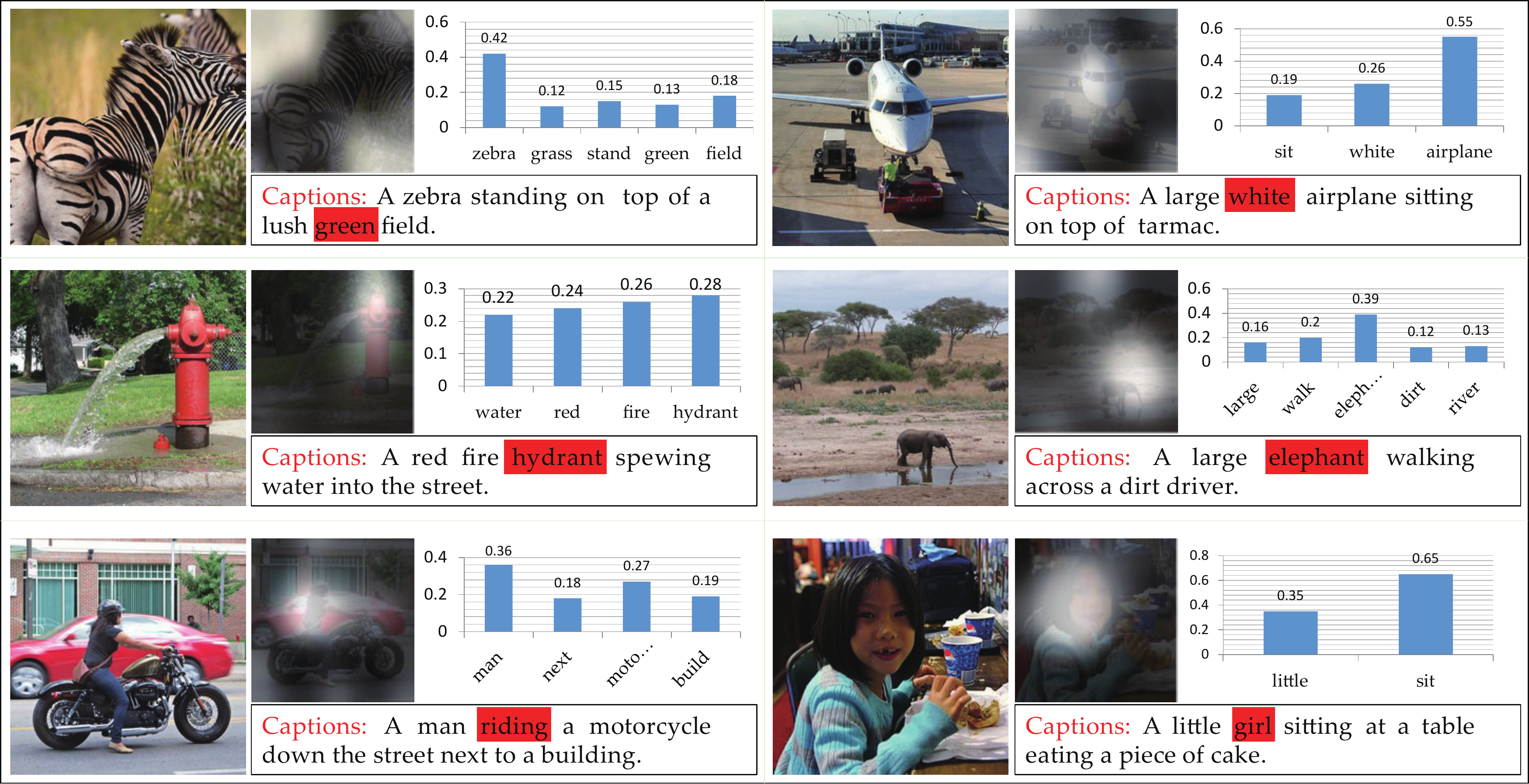}\\
  \caption{Visualization of generated captions, attributes and image attention maps on the MS COCO dataset. ``Captions'' denotes the describing sentences generated by our model. White areas of the attention maps indicate the attended regions which correspond to the captions. The histograms show the attention weights of the attributes. The related words are marked with red rectangular box (Best view in color).}
  \label{visualization}
\end{figure*}

\begin{table}[!htb]
\renewcommand{\arraystretch}{1}
\caption{Experimental results on MS COCO.}
\label{results_on _MSCOCO}
\begin{center}
\resizebox{\linewidth}{!}{
\small\begin{tabular}{c||ccccc}
  \thickhline
  Model & B-1 & B-2 & B-3 & B-4 & METEOR\\
  \hline\hline
  NeuralTalk \cite{DBLP:conf/cvpr/KarpathyL15}  & 62.5 & 45.0 & 32.1 & 23.0 & 19.5\\
  Goole-NIC \cite{DBLP:conf/cvpr/VinyalsTBE15} & - & - & - & 27.7 & 23.7\\
  LRCN \cite{DBLP:conf/cvpr/DonahueHGRVDS15} & 62.8 & 44.2 & 30.4 & 21.0 & -\\
  m-RNN \cite{DBLP:journals/corr/MaoXYWY14a} & 67 & 49 & 35 & 25 & -\\
  \hline\hline
  Att-SVM + LSTM \cite{DBLP:conf/cvpr/WuSLDH16} & 69 & 52 & 38 & 28 & 23\\
  Att-CNN + LSTM \cite{DBLP:conf/cvpr/WuSLDH16} & 74 & 56 & 42 & 31 & 26\\
  \hline\hline
  NIC-VA-Soft \cite{DBLP:conf/icml/XuBKCCSZB15} & 70.7 & 49.2 & 34.4 & 24.3 & 23.9\\
  NIC-VA-Hard \cite{DBLP:conf/icml/XuBKCCSZB15} & 71.8 & 50.4 & 35.7 & 25.0 & 23.0\\
  ATT-$k$-NN \cite{DBLP:conf/cvpr/YouJWFL16} & 67.6 & 50.5 & 37.5 & 28.1 & 22.7\\
  ATT-RK \cite{DBLP:conf/cvpr/YouJWFL16} & 67.9 & 50.6 & 37.5 & 28.2 & 23.1\\
  ATT-FCN \cite{DBLP:conf/cvpr/YouJWFL16} & 70.9 & 53.7 & 40.2 & 30.4 & 24.3\\
  RA \cite{Kun_Fu} & 71.7 & 54.8 & 40.9 & 30.2 & 24.2\\
  SS \cite{Kun_Fu} & 71.2 & 53.6 & 39.4 & 28.9 & 24.1\\
  RA+SS \cite{Kun_Fu} & 71.7 & 54.9 & 41.1 & 30.6 & 24.5\\
  (RA+SS)-ENSEMBLE \cite{Kun_Fu} & 72.4 & 55.5 & 41.8 & 31.3 & 24.8\\
  \hline\hline
  Ours & 73.9 & 56.4 & 41.7 & 30.9 & 27.1\\
  \thickhline
\end{tabular}}
\end{center}
\end{table}

Table \ref{results_on _MSCOCO} shows image captioning results on the MS COCO dataset. Similar to the experiment on the Flickr8K and Flickr30K, the contrast models also be classified into three categories, \emph{i.e.}, none attention and none attribute models (such as NeuralTalk, Google-NIC, LRCN and m-RNN), only attribute-based models (such as Att-CNN + LSTM) and only attention-based models (such as NIC-VA and ATT-FCN). Among the contrast models, Att-CNN +LSTM gets the highest scores both on BLEU and METEOR metrics. It shows that the high-level semantic information is important to transform an image into natural language sentence. That is to say, the high-level semantic information can contribute significantly to eliminate the semantic gap between vision and language. Compared with the proposed model in this paper, no attention mechanism has been used in Att-CNN + LSTM. In other words, the attributes information is encoded into one vector and imported into the language model. Table \ref{results_on _MSCOCO} shows that our model gets higher scores on most of the metrics. It implies that our model with attention mechanism is more effective than Att. In addition, the results of attention-based models are obviously better than the none attribute-based and attention-based models. The fact indicates that the attention mechanism can find fine-grained relationship between image region and sentence element, and this relationship is effective for image captioning.
\par Although the attention-based and attribute-based models show the powerful ability on image captioning task, our model further improves the performance. There are two main reasons: 1) the word-guided attention network can find the fine-grained relationship between image regions and words; 2) the semantic-guided attention network adds high-level semantic information which contributes to eliminating the semantic gap between vision and language.
\par Fig. \ref{visualization} shows the visualization of generated captions, attributes and image attention maps on the MS COCO dataset. According to the Fig. \ref{visualization}, we see that our model successfully learns to align the local image regions, image attributes and words. For instance, when generate captions for the first image in the third row, the attribute layer predicts four attributes (i.e., ``\emph{man}'', ``\emph{next}'', ``\emph{motorcycle}'' and ``\emph{build}'') of this image. When generate the word ``riding'', the proposed model attends the most related region (i.e., the man's region of the image). The histogram shows the attention weights of the four attributes, and the attention weights are computed by Eq.  (\ref{v}).  All the instances prove that the model can explore the relationships among the attributes, local image regions and captions very well.
\subsubsection{Ablation Study}
\begin{figure*}[t]
  \centering
  \includegraphics[width=0.98\linewidth]{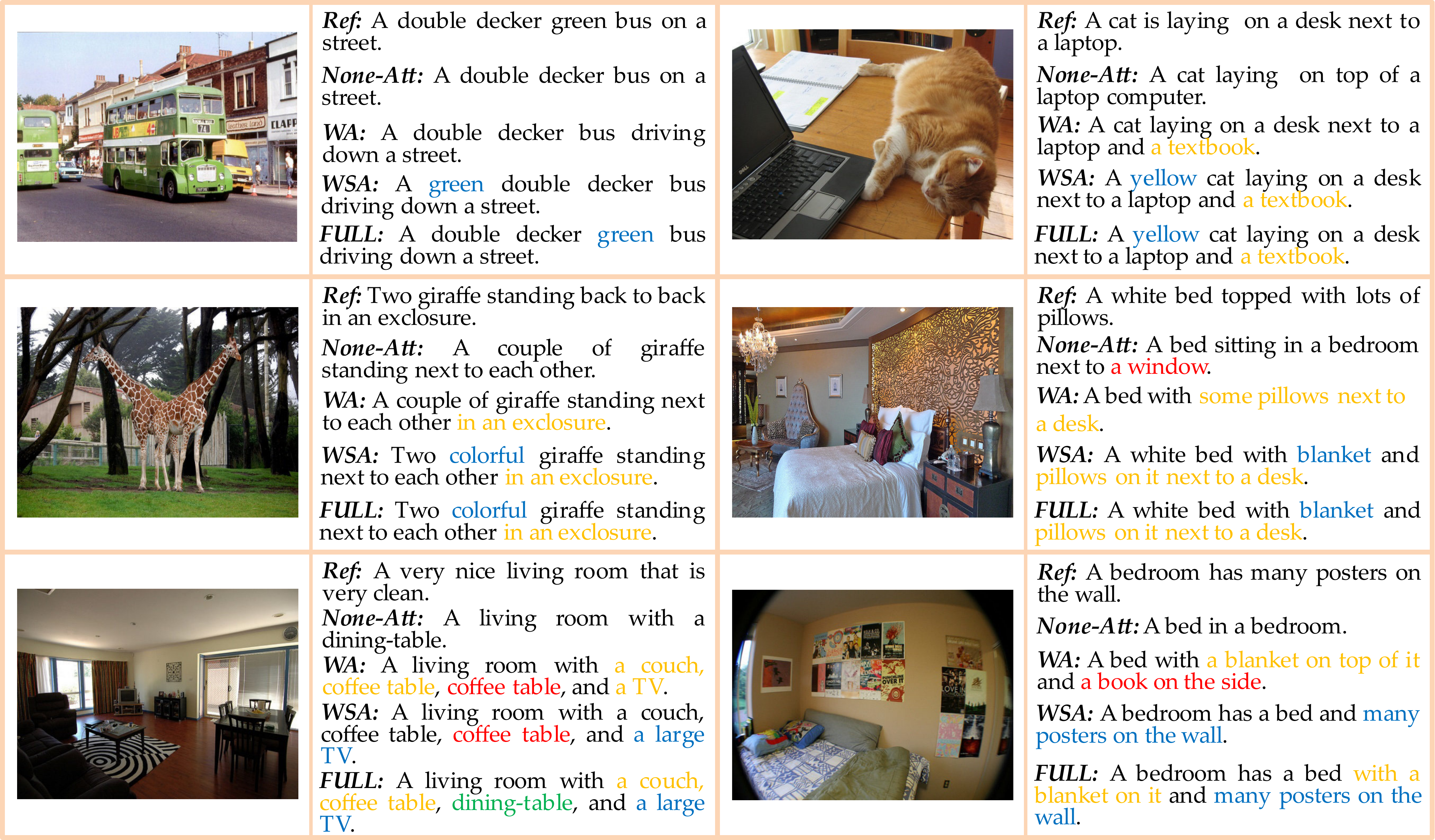}\\
  \caption{Some typical examples on the MS COCO validation subset. The Ref sentences are generated by humans, and the other sentences are generated by the ablation model. The wrong words generated by all models are signed with red font. The right generated words with different color are generated by the ablation models: the words with yellow font denotes the caption are not appeared in the Non-att model but generated by the models with visual attention mechanism; the blue font words are generated by the models with semantic-guided attention and the green words are only generated by our full model. Best view in color.}
  \label{example}
\end{figure*}

To verify the effectiveness of each component in our model, we perform ablation studies by ablating certain components:
\begin{itemize}
  \item None attention is used for image captioning (None-Att). The word-guided and semantic-guided attention networks are abandoned. Only the global image feature output from FC7 layer of VGG-19 are used to generate descriptions.
  \item Word-guided attention (WA). Only the word-guided attention network is used.
  \item Word and semantic-guided attention (WSA). The two attention networks are both used while the gate for $v'_I$ are abandoned.
  \item Word and semantic-guided attention with gating controlling (FULL). Our full model.
\end{itemize}
\begin{table}[!htb]
\caption{Ablation study on the MS COCO}
\label{basline_1}
\begin{center}
\small\begin{tabular}{l||ccccc}
  \thickhline
  Model & B-1 & B-2 & B-3 & B-4 & METEOR\\
  \hline\hline
  None-Att & 63.1 & 46.3 & 31.9 & 23.1 & 20.2\\
  WA & 70.8 & 50.2 & 37.4 & 25.8 & 23.8\\
  WSA & 72.9 & 53.4 & 41.2 & 30.7 & 27.9\\
  \hline\hline
  FULL &73.9 & 56.4 & 41.7 & 30.9 & 27.1\\
  \thickhline
\end{tabular}
\end{center}
\end{table}
Table \ref{basline_1} shows the performance of the ablation models. The results confirm the truth that both the word-guided and semantic-guided attention networks. First, the WA model improves the performance on the bias of the Non-Att model. That is because the word-guided attention network can automatically focus on the most word-related regions. Second, due to introducing the concept information and semantic-guided attention network, the WSA model further improves the performance on the bias of the WA model. The concept information is an important supplement for image information and the semantic-guided attention network explores the relationship between the concepts and regions. Last, our full model, which adds a gate to control the vector output from the semantic-guided attention model, is an improved version of the WSA model. As can be seen in Tabel \ref{basline_1}, the gate is necessary, because the gate automatically controls whether and how much the concept-region representation should be input to the RNN module at each time-step.
\par Fig. \ref{example} shows some examples of image captioning on the validation set of the MS COCO dataset. Generally speaking, our full model shows  best performance among the ablation models. The None-Att model may loss the attribute information and some important object information when generates description for image. For example, the second instance of the first row in Fig. \ref{example}, the None-Att model can describe the main object in the scene (such as ``\textsl{cat }'' and ``\textsl{laptop}''), but the attribute of the ``\textsl{cat}'' (\textsl{i.e.}, the color of the cat---yellow) and some important objects (such as ``\textsl{desk}'' and ``\textsl{textbook}'') are not been described. Two main reasons may  led to such a result: 1) the attribute information (\textsl{i.e.}, the color of the cat)  is not be used to generate the caption; and 2) the None-Att model does not have the word-guided attention network which can exploit the attention-transfer mechanism. In other words, the None-Att model only focuses on the main region (the ``\textsl{cat}'' and ``\textsl{laptop}'' region) and describes it, but it cannot transfer the attention into other regions (such as the ``\textsl{desk}'' and ``\textsl{textbook}'' region). The WA model tries to dig all important object in an image, but it may make some mistakes. For instance,  no ``\textsl{book}'' in the second image of the third row in Fig. \ref{example}, but the WA model identifies some object as ``\textsl{book}''. Simultaneously, the ``\textsl{posters}'' are missed. However, the WSA model has expressed the information of the ``\textsl{posters}''. This is mainly because the ``\textsl{poster}'' is an attribute word and the WSA has the semantic-guided attention network which can make full use of the attribute information. Our FULL model not only considers both the semantic information and the relationship between the word and image region, but also uses an gate to control when an how much the semantic information output from the semantic-guided attention network should be used to generated the description. This structure can correct some mistakes by the WSA. For example, when the WSA model describe the first image of the third row in Fig. \ref{example}, two ``\textsl{coffee table}'' are generated, but our FULL model correct this mistake and generate the right caption---``\textsl{dining-room}''.
\subsection{Visual Question Answering}

\begin{table*}[t]
\caption{Experimental results on COCO-QA.}
\label{results 2}
\begin{center}
\small\begin{tabular}{c||cccc|ccc}
  \thickhline
  Model & Object & Number & Color & Location &Accuracy & WUPS@0.9 & WUPS@0.0 \\
  \hline \hline
  VIS+BOW \cite{DBLP:conf/nips/RenKZ15} &60.17 &43.99 &52.97 &51.53 & 57.27 & 68.03 & 89.58 \\
  VIS+LSTM \cite{DBLP:conf/nips/RenKZ15} &56.54 &45.12 &45.58 &45.76 & 53.21 & 63.82 & 88.31 \\
  2-VIS+BLSTM \cite{DBLP:conf/nips/RenKZ15} &58.49 &44.57 &50.76 &47.70 & 55.53 & 65.88 & 88.91 \\
  DPP-Net \cite{DBLP:conf/cvpr/NohSH16} &- &- &- &- & 61.19 & 70.84 & 90.61\\
  \hline \hline
  Att-LSTM \cite{DBLP:conf/cvpr/WuSLDH16} &63.92 &51.83 &57.29 &54.84 & 61.38 & 71.15 & 91.58\\
  \hline\hline
  SAN \cite{DBLP:conf/cvpr/YangHGDS16} &64.5 &48.6 &57.9 &54.0 & 61.6 & 71.6 & 90.9 \\
  ABC-CNN \cite{DBLP:journals/corr/ChenWCGXN15} &62.46 &45.7 &46.81 &53.67 &58.10 &68.44 & 89.85\\
  CoATT + VGG \cite{DBLP:conf/nips/LuYBP16} &65.6 &49.6 &61.5 &56.8 &63.3 &73.0 & 91.3\\
  \hline\hline
  Ours &67.51 &51.55 &62.10 &56.78 & 64.32 & 74.89 & 92.02\\
   \thickhline
\end{tabular}
\end{center}
\end{table*}

\begin{table*}[t]
\caption{Experimental results on VQA.}
\label{results on VQA}
\newcommand{\tabincell}[2]{\begin{tabular}{@{}#1@{}}#2\end{tabular}}
\begin{center}
\small\begin{tabular}{c||cccc|cccc}
 \thickhline
  &\multicolumn{4}{c|}{Test-dev}  &\multicolumn{4}{c}{Test-standard} \\
   \cline{2-9}\raisebox{0.5em}{Model} &\tabincell{c}{Yes/No} & \tabincell{c}{Number} & \tabincell{c}{Others} & \tabincell{c}{All} &\tabincell{c}{Yes/No} & \tabincell{c}{Number} & \tabincell{c}{Others} & \tabincell{c}{All}\\
  \hline \hline
  LSTM Q + I  \cite{DBLP:conf/iccv/AntolALMBZP15} & 78.9 & 35.2 & 36.4 & 53.7 & 79.0 & 35.6 & 36.8 & 54.1\\
  DPP-Net \cite{DBLP:conf/cvpr/NohSH16} & 80.7 & 37.2 & 41.7 & 57.2 & 80.3 & 36.9 & 42.2 & 57.4 \\
  \hline \hline
  SAN \cite{DBLP:conf/cvpr/YangHGDS16} & 79.3 & 36.6 & 46.1 & 58.7 & - & - & - & 58.9\\
  CoATT \cite{DBLP:conf/nips/LuYBP16} & 79.5 & 38.7 & 48.3 & 60.1 & - & - & - & -\\
  SMem-VQA Two-Hop \cite{DBLP:conf/eccv/XuS16} & 80.87 & 37.32 & 43.12 & 57.99 & 80.8 & 37.53 & 43.48 & 58.24\\
  DAN (VGG) \cite{DBLP:conf/cvpr/NamHK17} & 82.1 & 38.2 & 50.2 & 62.0 & - & - & - & - \\
  DAN (ResNet) \cite{DBLP:conf/cvpr/NamHK17} & 83.0 & 39.1 & 53.9 & 64.3 & 82.8 & 38.1 & 54.0 & 64.2 \\
  \hline\hline
  Att-LSTM \cite{DBLP:conf/cvpr/WuSLDH16} & 78.90 & 36.11 & 40.07 & 55.57 & 78.73 & 36.08 & 40.60 & 55.84\\
  \hline\hline
  MLAN (ResNet) \cite{MLAN} & 82.9 & 39.2 & 52.8 & 63.7 & - & - & - & -\\
  MLAN (ResNet, train + val) \cite{MLAN} & 83.8 & 40.2 & 53.7 & 64.6 & 83.7 & 40.9 & 53.7 & 64.8\\
  MLAN (ResNet, train + val + VG) \cite{MLAN} & 81.8 & 41.2 & 56.7 & 65.3 & 81.3 & 41.9 & 56.5 & 65.2\\
  \hline\hline
  Ours (VGG) & 83.5 & 41.5 & 55.8 & 65.2 & 83.2 & 42.0 & 55.1 & 65.5\\
  Ours (ResNet) & 83.7 &41.3 & 56.5 & 66.0 & 83.4 & 42.1 & 56.0 & 65.8\\
 \thickhline
\end{tabular}
\end{center}
\end{table*}
\subsubsection{Dataset and Evaluation Metrics}
\textbf{Dataset.} We report VQA results on Toronto COCO-QA, VQA dataset which are most popular publicly available visual question answering datasets based on MS COCO. Toronto COCO-QA dataset contains 8,000 images with 79,000 question/answer pairs for training and 4,000 images with 39,171 question/answer pairs for testing. The questions have four types (\emph{i.e.}, object, number, color and location). The answers are all single-word.
 VQA dataset is a much larger dataset which contains 614,163 questions. The training and testing split follows COCO official split, which contains 82,783 training images, 40,504 validation images and 81,434 test images, each has 3 questions and 10 answers. We use the official test split for our testing. The dataset has two different tasks : open-ended and multiple-choice tasks. We only report the experiment result on open-ended task.
\par \textbf{Evaluation Metrics.} We formulate VQA as a classification problem. The proposed model is evaluated with classification accuracy. The WUPS score \cite {wu1994verbs} is also reported. The WUPS calculates the similarity between two words based on the similarity between their common subsequence in the taxonomy.
\subsubsection{Results on COCO-QA Dataset}
Table \ref{results 2} shows the results on the COCO-QA dataset. We categorize the contrast models as \rmnum{1}) none attribute and attention models, \rmnum{2}) only attribute-based models and \rmnum{3}) only attention-based models (every category is separated with double horizontal line in Table \ref{results 2}). From the Table \ref{results 2}, we can easily draw a conclusion that both the attention-based models and the attribute-based models significantly improved accuracy (about $5\%$  increase on both the four types of questions) on the COCO-QA dataset. The Att-LSTM model shows more powerful performance on the COCO-QA dataset than the none-att based (none-attribute and none-attention based) models and the attention-based models (except the CoATT + VGG model \cite{DBLP:conf/nips/LuYBP16} which include image attention and question attention). It confirms that the high-level semantic information is important to solve the VQA problem. In addition, the results of attention-based models are obviously better than the none att-based and attention-based models. It main because the attention model can focus on the important region which is very correlation with the question.
\par Through the results in Table \ref{results 2}, we find that our model improves the state-of-the-art from $63.3\%$ (CoATT + VGG \cite{DBLP:conf/nips/LuYBP16}) to $64.32\%$. For the different types of questions, all the models in Table \ref{results 2} show less powerful performance on the Number and Location questions than the Object and Color Question. That mainly caused by unbalanced data: the COCO-QA dataset contains $70\%$ Object questions, $7\%$ Number questions, $17\%$ Color questions and $6\%$ Location questions. However, our model and the Att-LSTM model increase the accuracy much greater than the attention-based models on the Number questions. Furthermore, the proposed method is further improved the performance for the Object questions. Two main reasons cause the result: 1) the question-guided attention network focuses on the important region which relates to the question; 2) the semantic-guided-attention provide the attribute information and the object noun is the main element for the attribute set. All the results show that the proposed method outperforms almost all the contrast model on all types of questions.

\subsubsection{Results on VQA Dataset}
\begin{figure*}[ht]
  \centering
  \includegraphics[width=0.98\linewidth]{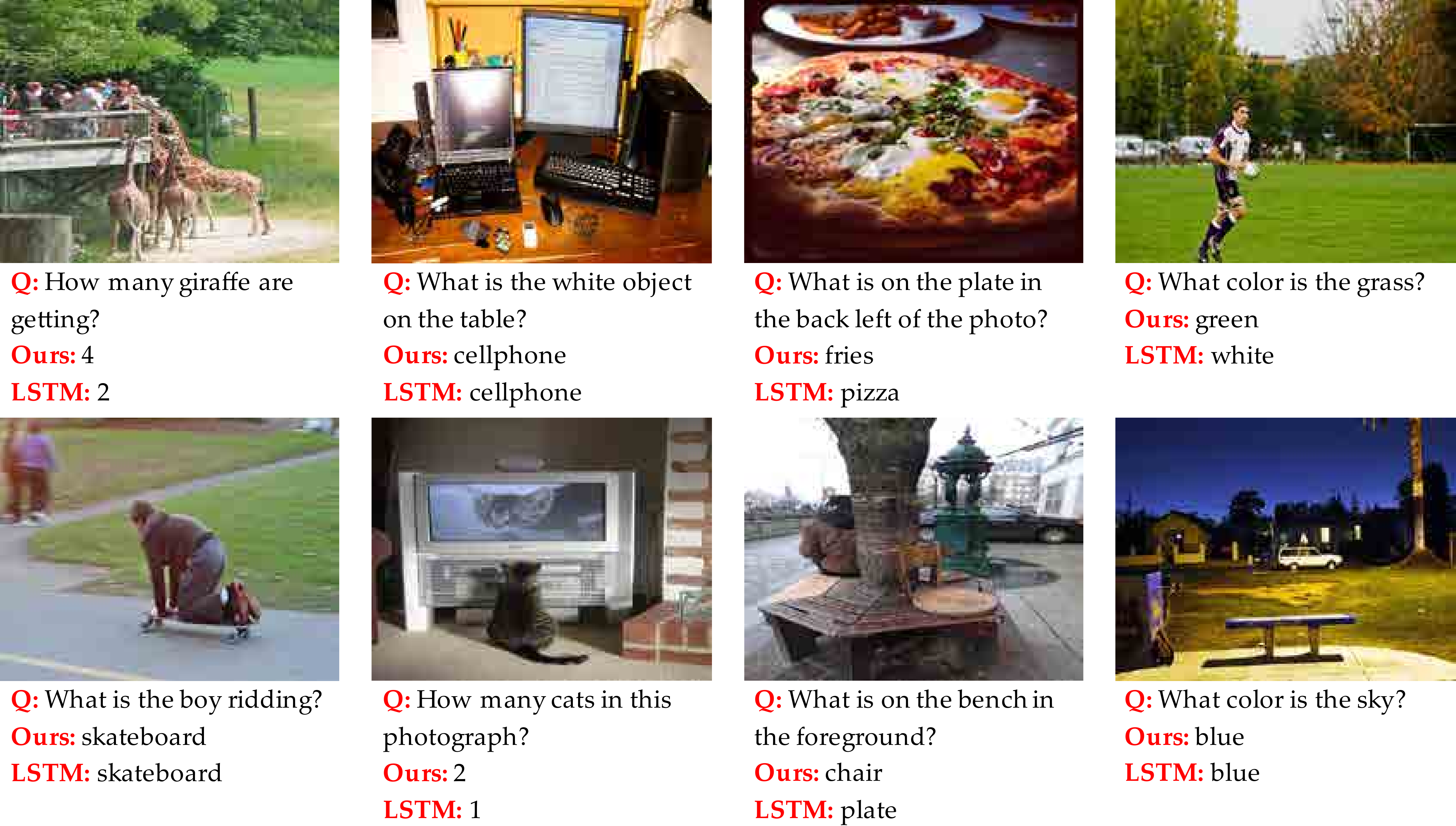}\\
  \caption{Some typical examples on the VQA validation subset. The letter Q denotes the corresponding question, ``Ours'' and ``LSTM'' represent the answers generated by our model and the VIS+LSTM proposed in reference \cite{DBLP:conf/nips/RenKZ15}, respectively. Considering that VIS+LSTM is a typical model that uses the output of the penultimate full connection layer of the CNN as the image representation. Furthermore, the question is encoded into a vector by LSTM. And both the image representation and the question vector are input into the softmax layer to generate the answer.}
  \label{vqa_example}
\end{figure*}
We compare our method with several state-of-the-art methods on the VQA dataset. For equality, the per-answer category accuracy and overall accuracy of the models are shown in Table \ref{results on VQA}. The compared models are divided into four categories: \rmnum{1}) none attribute and attention models, \rmnum{2}) only attribute-based models, \rmnum{3}) only attention-based models and \rmnum{4}) both the semantic-guided attention and attention-guided attention models (every category is separated with double horizontal line in Table \ref{results on VQA}). It is observed that our model get almost the best performance on all category questions. Only the MLAN model \cite{MLAN} achieves a comparable results with ours. Two reasons for such a result. Firstly, the MLAN model uses two level attention networks, the visual attention which is benefit for fine-grained spatial inference and the semantic attention which reduces the semantic gap.  Secondly, the MLAN model use the ResNet \cite{DBLP:conf/cvpr/HeZRS16} as image information extractor, which is more powerful than the VGG-Net. However, our model is not inferior to it. To get a more fair comparison (\emph{i.e.}, to eliminate the influence of the visual features), ResNet-152 is also used as image feature extractor (Ours (ResNet) in Table \ref{results on VQA}). Specially, feature maps output from the last convolutional layer of the ResNet-152 are used as visual features. So each image is cropped into 49 regions and each region is represented as a 2048-dimension vector. Compared with MLAN, almost for all types of questions our gets the highest scores. It shows that with the same image feature representation, our model shows a better performance than MLAN. Furthermore, there are two differences between MLAN and ours: (a) our approach is designed for both image captioning and visual questioning answering, while MLAN only used for VQA; (b) the semantic-guided attention in our approach is a dual structure, \emph{i.e.}, the semantic-guided attention network is used not only to find the most attribute-related image regions, but also to find the most region-related attributes. While the semantic attention network in MLAN is used to find the attributes coressponding to the question. The experimental results shown in Table \ref{results on VQA} demonstrates that the dual structure of the semantic-guided attention is efficiency for VQA task. All results in Table \ref{results 2} and \ref{results on VQA} confirmed that the semantic information and the attention network are vital to VQA task.
\par Fig. \ref{vqa_example} shows some typical examples on the VQA validation subset. The letter Q denotes the corresponding questions, ``Ours'' and ``LSTM'' represent the answers generated by our model and the VIS+LSTM proposed in reference \cite{DBLP:conf/nips/RenKZ15}, respectively. According to the instances shown in Fig. \ref{vqa_example}, we can find that our model shows a much better performance than the VIS+LSTM, especially on the number questions. For example, the second instance of the second row, two cats are in the photo, but the VIS+LSTM only finds one. It mainly because the proposed model contains two level attention networks which helps the model focus on the most related regions for the question. Moreover, when answering the object attribute’s question, the proposed model gets more accurate answers than VIS + LSTM. For example, the forth image in the first row show a big football field. Obviously, the football field is green, while the VIS+LSTM judges the grass is white. The examples show that our model is effective for the VQA task.

\begin{table*}[ht]
\caption{Ablation study on COCO-QA.}
\label{basline_2}
\begin{center}
\small\begin{tabular}{c||cccc|ccc}
  \thickhline
  Model & Object & Number & Color & Location &Accuracy & WUPS@0.9 & WUPS@0.0 \\
  \hline\hline
  None-Att &56.52 &45.22 &48.35 &46.24 & 53.72 & 64.02 & 88.51 \\
  QA &64.27 &49.13 &57.72 &54.33 & 61.50 &70.87 & 90.43 \\
  SA &64.38 &51.22 &59.24 &54.86 & 62.01 & 71.42 & 91.34 \\
  \hline\hline
  FULL &67.51 &51.55 &62.10 &56.78 & 64.32 & 74.89 & 92.02 \\
  \thickhline
\end{tabular}
\end{center}
\end{table*}
\subsubsection{Ablation Study}
To further verify the effectiveness of each component in our model, we perform ablation studies by ablating certain components:
\begin{itemize}
  \item None-attention (None-Att). Only the image representation $v_I$ and question representation $v_q$ are used for VQA problem ,without the semantic-guided attention network and question-guided attention network.
  \item Question-guided attention (QA). Only the question-guided attention network are used.
  \item Semantic-guided attention (SA). Only the semantic-guided attention network are used.
  \item Question and semantic-guided attention (FULL). All attention networks are used for VQA.
\end{itemize}

\par Table \ref{basline_2} shows the performance of the ablation models on the COCO-QA dataset. The results are similar to those in Table \ref{basline_1}, so we can get a similar conclusion: 1) the question-guided attention network is effective in finding question-related regions and 2) the semantic-guided attention network provide more important supplement information which helps the model get the correct answers.
\section{Conclusion} \label{Conclusion}
We propose a novel model based on attributes and attention mechanism for V2L problems. The model concludes two level attention networks. The text-guided attention network enables subtle understanding between vision and language, and the semantic-guided attention network provides high-level concepts information and explores the subtle relationships between concepts and regions which reduces the gap between language and visual information. Our model makes full use of the complementarity of the different level visual representations.
The extensive experiments both for image captioning and visual question answering show that our model outperforms any single visual attention or attribute model. The semantic attention network is an important supplement for text-guided attention.

\ifCLASSOPTIONcaptionsoff
  \newpage
\fi

\bibliographystyle{IEEEtran}
\bibliography{egbib}

\begin{IEEEbiographynophoto}{Xuelong Li}
is a full professor with School of Computer Science and Center for OPTical IMagery Analysis and Learning (OPTIMAL), Northwestern Polytechnical University, Xi'an 710072, P.R. China.
\end{IEEEbiographynophoto}

\begin{IEEEbiography}[{\includegraphics[width=1in,height=1.25in,clip,keepaspectratio]{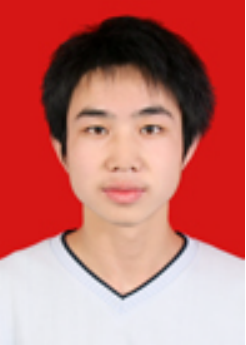}}]{Aihong Yuan}
is currently pursuing the Ph.D. degree with the Key Laboratory of Spectral Imaging Technology CAS, Xi'an Institute of Optics and Precision Mechanics, Chinese Academy of Sciences, Xi'an 710119, Shaanxi, P. R. China. His research interests include image/video content understanding and deep learning.
\end{IEEEbiography}

\begin{IEEEbiography}[{\includegraphics[width=1in,height=1.25in,clip,keepaspectratio]{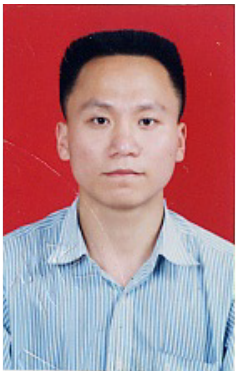}}]{Xiaoqiang Lu}
is a Full Professor with the Key Laboratory of Spectral Imaging Technology CAS, Xi'an Institute of Optics and Precision Mechanics, Chinese Academy of Sciences, Xi'an 710119, Shaanxi, P. R. China.
\end{IEEEbiography}

\end{document}